\documentclass[lettersize,journal]{IEEEtran}
\usepackage{amsmath,amsfonts}
\usepackage{algorithmic}
\usepackage{algorithm}
\usepackage{array}
\usepackage{textcomp}
\usepackage{stfloats}
\usepackage{url}
\usepackage{verbatim}
\usepackage{graphicx}
\usepackage{cite}
\usepackage{amssymb}
\usepackage{makecell}
\usepackage{amsmath}
\usepackage{multirow}
\usepackage{listings}
\usepackage{color, colortbl}
\usepackage{xcolor}
\usepackage{circledtext}
\usepackage{booktabs}
\usepackage{lineno}
\usepackage{hyperref}
\definecolor{codegreen}{rgb}{0,0.6,0}
\definecolor{red}{rgb}{0.8,0,0}
\definecolor{blue}{rgb}{0,0,0.8}
\definecolor{Gray}{rgb}{0.9,0.9,0.9}
\usepackage[caption=false,font=small]{subfig}

\newcolumntype{L}[1]{>{\raggedright\let\newline\\\arraybackslash\hspace{0pt}}m{#1}}
\newcolumntype{C}[1]{>{\centering\let\newline\\\arraybackslash\hspace{0pt}}m{#1}}
\newcolumntype{R}[1]{>{\raggedleft\let\newline\\\arraybackslash\hspace{0pt}}m{#1}}

\usepackage{tikz}

\begin{document}

\title{EV-CLIP: Efficient Visual Prompt Adaptation for CLIP in Few-shot Action Recognition under Visual Challenges}

\author{Hyo Jin Jon$^*$, Longbin Jin$^*$, Eun Yi Kim
\thanks{H. J. Jon, L. Jin, and E. Y. Kim are with Artificial Intelligence \& Computer Vision Lab., Konkuk University, Seoul, South Korea (e-mail: hyojin2011@konkuk.ac.kr; jinlongbin@konkuk.ac.kr; eykim@konkuk.ac.kr).}
\thanks{$*$ denotes equal contribution.}
}



\maketitle

\begin{abstract}
CLIP has demonstrated strong generalization in visual domains through natural language supervision, even for video action recognition. However, most existing approaches that adapt CLIP for action recognition have primarily focused on temporal modeling, often overlooking spatial perception. In real-world scenarios, visual challenges such as low-light environments or egocentric viewpoints can severely impair spatial understanding, an essential precursor for effective temporal reasoning. To address this limitation, we propose Efficient Visual Prompting for CLIP (EV-CLIP), an efficient adaptation framework designed for few-shot video action recognition across diverse scenes and viewpoints. EV-CLIP introduces two visual prompts: mask prompts, which guide the model’s attention to action-relevant regions by reweighting pixels, and context prompts, which perform lightweight temporal modeling by compressing frame-wise features into a compact representation. For a comprehensive evaluation, we curate five benchmark datasets and analyze domain shifts to quantify the influence of diverse visual and semantic factors on action recognition. Experimental results demonstrate that EV-CLIP outperforms existing parameter-efficient methods in overall performance. Moreover, its efficiency remains independent of the backbone scale, making it well-suited for deployment in real-world, resource-constrained scenarios. The code is available at \href{https://github.com/AI-CV-Lab/EV-CLIP}{https://github.com/AI-CV-Lab/EV-CLIP}.
\end{abstract}


\section{Introduction}
\IEEEPARstart{R}{ecognizing} human actions is a fundamental step toward understanding human behavior, with applications in robotics \cite{robotics}, surveillance systems \cite{surveillance}, and wearable devices such as smart glasses \cite{smartglasses}. Over the past decade, deep neural networks, such as convolutional neural networks \cite{c3d, p3d, i3d, r2+1d, nonlocalnet, lin2019tsm, tea, tdn} and transformers \cite{vivit, timesformer, vtn, vst, mtv, li2022uniformer}, have driven significant progress in recognizing actions from video frames. However, in real world, videos introduce diverse visual challenges, including variations in viewpoints and illuminations. Unlike clean lab-based video data, these visual challenges may degrade the action recognition of models. For example, egocentric viewpoints may obscure the action performer, and low-light conditions hinder object visibility.

While training large-scale video models to handle such diverse visual conditions is theoretically feasible, it remains impractical in real-world deployments due to substantial data volume, annotation, and computational resources requirements. Domain adaptation has emerged as a more scalable alternative by transferring knowledge from pretrained models to new target domains, reducing the need for retraining from scratch. However, most existing approaches, such as unsupervised domain adaptation (UDA) \cite{ta3n, shuffleandattend, tcon, mmsada, amls, ptc, bds, stcda, cmco, comix, co2a} and few-shot domain adaptation (FSDA) \cite{pastn, fs-ada, ssa2lign}, still assume either access to large volumes of target data or the availability of source datasets, which may be inaccessible in real-world scenarios. These assumptions often break down in practical settings, especially when models are accessed as black-box APIs or when their training datasets are proprietary. As foundation models increasingly rely on web-scale, privately held datasets, the need for source-free few-shot adaptation becomes more critical.

Visual-language models (VLMs) such as CLIP \cite{clip} have demonstrated remarkable generalization capabilities in vision tasks by embedding images and text into a shared semantic space. This language supervision enables VLMs to recognize novel categories and adapt across domains without requiring access to the original training data. Leveraging this strength, recent efforts have explored adapting CLIP to video action recognition through additional learning schemes \cite{vifi-clip, x-clip, actionclip, claver, maxi, ost, dist, evl, froster, dual-path, aim, st-adapter, vop, vilt-clip, vita-clip, a5, ez-clip, open-vclip, open-vclip++}. While zero-shot approaches depend on extensive post-pretraining on large-scale video datasets such as Kinetics-400 \cite{kinetics}, few-shot approaches offer a more practical alternative by transferring knowledge using only a small number of labeled examples. This aligns more closely with real-world settings, where collecting large-scale video datasets is costly, but annotating a small subset is feasible. In this work, we investigate how to adapt CLIP under such few-shot constraints to recognize actions in visually challenging conditions, including low-light environments and egocentric viewpoints.

\begin{figure*}[!t]
\centering
\subfloat[]{
    \includegraphics[width=0.28\textwidth]{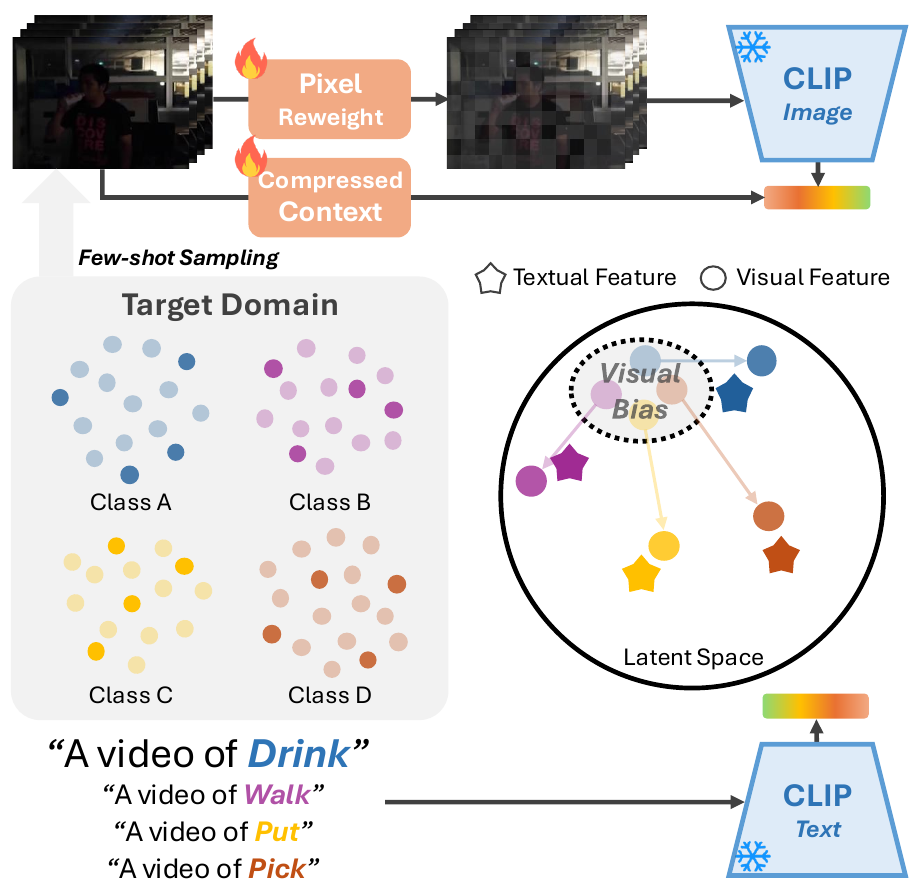}
    \label{fig;main1}
}
\hspace{0.05\textwidth}
\subfloat[]{
    \includegraphics[width=0.56\textwidth]{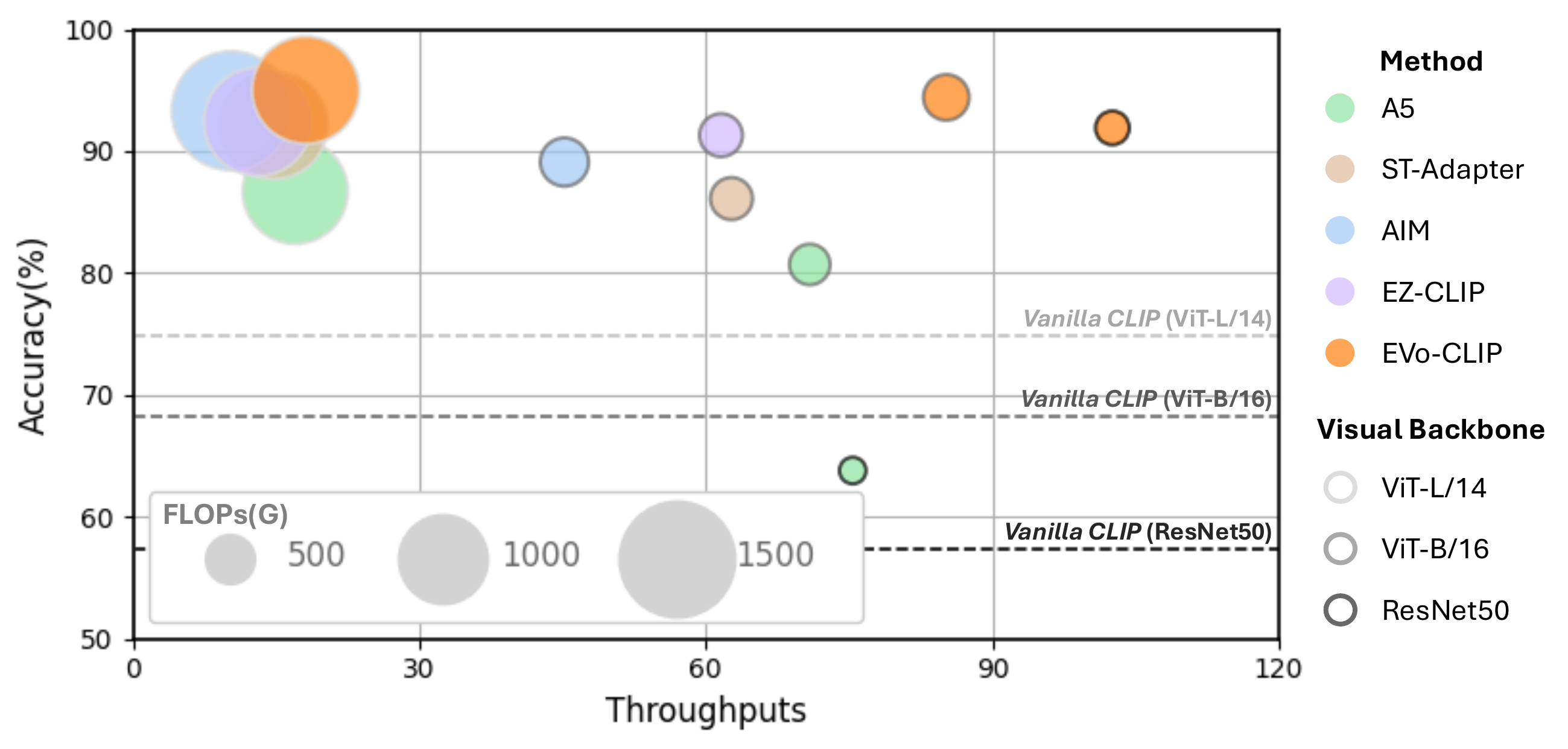}
    \label{fig;main2}
}
\caption{(a) Main approach of EV-CLIP. Few-shot samples from the target domain are adapted via two visual prompts, pixel reweighting for spatial focus and compressed context aggregation for temporal cues, aligning visual features with textual semantics in CLIP’s latent space. (b) Comparison of accuracy, throughput, and FLOPs on UCF101 under eight-shot settings. EV-CLIP achieves strong performance with efficiency across backbone choices.}
\label{fig;main}
\end{figure*}

Despite recent advances, existing CLIP-based action recognition methods have predominantly focused on temporal modeling using videos with clear visibility, where spatial understanding is already reliable. However, real-world scenarios often involve substantial visual challenges, such as low-light environments or egocentric viewpoints, that significantly hinder spatial perception. In such conditions, accurate action recognition requires robust spatial understanding before effective temporal reasoning can take place. Furthermore, current approaches frequently suffer from inefficiency, limited architectural generality, and scalability issues. Their performance is often tightly coupled with specific backbone choices, resulting in undesirable trade-offs between accuracy, trainable parameters, and throughput. These limitations pose challenges to practical deployment in diverse real-world settings.

To address these issues, we introduce Efficient Visual Prompt Adaptation for CLIP (EV-CLIP), a modular framework designed to enhance visual adaptability while preserving efficiency across diverse backbone encoders. As illustrated in Fig.~\ref{fig;main1}, EV-CLIP introduces two lightweight visual prompts that adapt the frozen CLIP visual encoder without altering its internal architecture. The mask prompt enhances spatial focus by selectively reweighting pixel intensities in input frames, emphasizing action-relevant regions and suppressing background noise. The context prompt performs lightweight temporal modeling by compressing frame-wise features into a single contextual vector, which is then pooled with individual frame features. Notably, both prompts are lightweight, backbone-agnostic, and compatible with CNN- and transformer-based CLIP encoders, enabling scalable deployment under varying computational constraints.

To ensure a comprehensive and diverse evaluation, we first curate and analyze five benchmark datasets, quantifying domain shifts from multiple perspectives. This analysis highlights the breadth of visual challenges, including low-light environments and egocentric viewpoints, encountered in real-world video scenarios. Evaluations across these benchmarks demonstrate that EV-CLIP consistently achieves the highest overall performance in few-shot adaptation settings, surpassing prior methods. As shown in Fig.~\ref{fig;main2}, EV-CLIP maintains strong accuracy even with lightweight backbones such as ResNet50, significantly reducing computational overhead without compromising recognition performance. These results underscore the practicality and scalability of EV-CLIP as an efficient solution for video domain adaptation in resource-constrained, real-world environments.

\section{Related Work}
\subsection{Video Action Recognition}
Traditionally, video action recognition has evolved by building upon advances in image-based models, with both CNNs and transformers emerging as dominant approaches. Early work employed 3D CNNs \cite{c3d, p3d, i3d} to jointly capture spatial and temporal features by extending 2D kernels into the temporal axis. Variants such as P3D \cite{p3d} incorporated residual connections, while I3D \cite{i3d} introduced additional modalities such as optical flow to enhance temporal modeling. To improve computational efficiency, many approaches decouple spatial and temporal learning by combining 2D CNN backbones with specialized temporal modules \cite{r2+1d, nonlocalnet, lin2019tsm, tea, tdn}, enabling more scalable training on large-scale video datasets.

More recently, the introduction of Vision Transformers (ViT) \cite{vit} has driven a paradigm shift toward transformer-based video models. ViViT \cite{vivit} extended ViT by generating spatiotemporal tubelets as input tokens, while TimeSformer \cite{timesformer} and VTN \cite{vtn} proposed factorized attention mechanisms that separately model spatial and temporal dependencies. Video-Swin \cite{vst} adapted the hierarchical Swin Transformer architecture \cite{swin} to the video domain, while Uniformer \cite{li2022uniformer} introduced a hybrid design that integrates 3D convolutions with transformers for unified spatiotemporal learning. Omnivore \cite{omni} further generalized this framework to process diverse input modalities, including images, videos, and RGB-D data, within a single unified model.

While these models have shown strong performance, retraining them from scratch for every new target domain is often impractical due to the substantial demands on data, annotation, and computation. To address this, domain adaptation has emerged as an effective and efficient alternative, enabling the transfer of knowledge from pretrained models to novel target domains.

\subsection{Domain Shift in Video Action Recognition}
In real-world video action recognition, models often encounter substantial domain shifts caused by variations in illumination, viewpoints, backgrounds, and camera perspectives. Such domain shifts can significantly degrade model performance when deployed outside of controlled training environments. Domain adaptation has become a key strategy to address these challenges by transferring knowledge from pretrained models to new target domains with differing data distributions.

UDA has been the most extensively studied paradigm for domain shift in video action recognition. UDA methods leverage labeled source data and unlabeled target data to learn domain-invariant representations. Broadly, adversarial-based approaches \cite{ta3n, shuffleandattend, tcon, mmsada} seek to disentangle domain-specific and content features via adversarial learning, discrepancy-based methods \cite{amls, ptc, bds} aim to minimize discrepancy between source and target feature distributions, and semantic-based approaches \cite{stcda, cmco, comix, co2a} aim to obtain domain-invariant features. More recently, source-free domain adaptation methods such as ATCoN \cite{atcon} have removed the dependency on source data entirely during adaptation. However, these approaches still typically require access to large volumes of unlabeled target data, limiting their practicality in scenarios where data collection is expensive or infeasible.

FSDA has emerged as a more practical solution for such data-scarce environments. FSDA adapts models using only a handful of labeled target samples per class, eliminating the need for abundant unlabeled data while reflecting real-world deployment conditions where limited annotation is feasible. In this setting, approaches such as PASTN \cite{pastn} and FS-ADA \cite{fs-ada} employ adversarial training to learn domain-invariant representations, while SSA$^2$lign \cite{ssa2lign} aligns source and target features at the snippet level via prototypical representations to improve few-shot transferability. However, many FSDA approaches still rely on access to source data, which is increasingly unavailable as foundation models are pretrained on massive, privately curated datasets. This growing restriction has amplified the demand for source-free FSDA techniques capable of effectively adapting models to new domains without requiring access to their original training data.

\subsection{Linguistic Supervision for Action Understanding}
The incorporation of linguistic supervision has emerged as a transformative paradigm across diverse visual tasks, including video anomaly detection \cite{anomize}, motion generation \cite{sato}, and action recognition. In particular, VLMs \cite{clip, align} have introduced significant generalization capabilities to vision tasks by leveraging linguistic supervision. By aligning visual and textual representations within a shared embedding space, they can recognize novel actions through their semantic understanding of action categories, even in previously unseen domains.

CLIP has become the most widely adopted VLM for video understanding. However, it was originally designed for static images, leading to modality gaps when applied directly to video data. To address this, numerous studies have sought to adapt CLIP for video action recognition by reducing the gap between images and temporal video sequences. ViFi-CLIP \cite{vifi-clip} fine-tunes CLIP by simply averaging frame-level features, while approaches such as X-CLIP \cite{x-clip}, ActionCLIP \cite{actionclip}, CLAVER \cite{claver}, MAXI \cite{maxi}, and OST \cite{ost} introduce additional modules to better model frame-wise dependencies. However, many of these methods update all parameters of CLIP along with newly added modules, resulting in high computational costs and limited scalability.

To mitigate these inefficiencies, parameter-efficient tuning methods (PETMs) have been proposed, aiming to adapt CLIP while keeping the majority of its pretrained parameters frozen. These efficient CLIP-based video learners incorporate lightweight trainable components, typically adapters or prompts, into the frozen backbone. Adapter-based approaches \cite{dist, evl, froster, dual-path, aim, st-adapter} insert small learnable modules between transformer layers to refine intermediate features. Prompt-based methods \cite{vop, vilt-clip, vita-clip} follow the VPT \cite{vpt} paradigm by inserting trainable prompt tokens into each transformer layer. A5 \cite{a5} introduces additional temporal heads for frame aggregation, while EZ-CLIP \cite{ez-clip} combines both adapters and prompts, and augments label semantics using GPT-generated textual descriptions.

Despite these advancements, existing PETMs have also predominantly focused on temporal modeling, often overlooking the importance of enhancing spatial perception at the frame level. Moreover, their architectures are typically tightly coupled with transformer-based backbones, and their efficiency tends to degrade significantly as backbone size increases. This scalability issue arises because adaptation modules introduce a growing number of trainable parameters with larger backbones, leading to increased computational costs that limit real-world applicability. In contrast, we propose EV-CLIP, which introduces modular visual prompts designed to improve spatial perception and support lightweight temporal modeling. Crucially, EV-CLIP maintains efficiency across diverse backbone architectures and scales, offering a flexible and deployment-friendly solution for practical video action recognition.

\section{Evaluation Protocol}
We establish a comprehensive evaluation protocol to benchmark CLIP-based action learners under few-shot learning scenarios, focusing on diverse real-world visual challenges. In this section, we first describe the CLIP-based framework for video action recognition, followed by a quantitative analysis of the domain characteristics across the evaluation datasets.

\subsection{Preliminary}
\noindent\textbf{Problem Formulation.}
Video action recognition aims to accurately identify human actions or interactions depicted within video sequences. Formally, let us define a set of domains as $\mathcal{D}\in\{\mathcal{D}_i\}_{i=1}^I$, where $I$ denotes the number of distinct domains. Within each domain $\mathcal{D}$, we have a set of video-label pairs, $\{(X_i, Y_i)\}_{i=1}^N$, where each video clip $X_i$ belongs to the space of all possible videos $\mathcal{X}$, and each corresponding label $Y_i$ belongs to the label space $\mathcal{Y}$. Specifically, a video clip represented as $X_i=\{x_{i, j}\}_{j=1}^T\in\mathbb{R}^{T\times C\times H\times W}$, where $T$ indicates the number of frames in the video sequence. The label texts $Y_i$ cover $M$ difference categories specific to the domain. 

The main objective in video action recognition is to correctly match each video clip with its corresponding textual action label, effectively increasing the model's ability to distinguish correct video-text pairs from incorrect ones. To reflect practical and real-world settings, where labeled data is typically limited, we explicitly emphasize data efficiency through few-shot learning scenarios. Specifically, each model learns actions from only $K$ labeled video clips per action category, enabling a realistic evaluation of performance under limited data conditions.

\noindent\textbf{CLIP-based Video Action Recognition.}\label{sec;CLIP-VL}
Originally designed for static images, CLIP leverages linguistic supervision to enhance visual understanding. Given its strong generalizability on image tasks, extending CLIP to video domains holds great potential for improving action understanding.

CLIP jointly employs visual and textual encoders, mapping images and text into a shared latent space via contrastive learning. Specifically, the textual encoder processes input tokens using a 12-layer transformer \cite{transformer}, while the visual encoder extracts image features using either ViT \cite{vit} or CNN-based residual networks (ResNet) \cite{resnet}. The resulting features from both modalities are projected into a unified latent embedding space using a linear layer.

Unlike the web-collected image-text pairs used during CLIP pretraining, video datasets typically lack informative captions. To mitigate this limitation, we adopt a simple text template for each action category. Thus, the textual representation is obtained by $t_i=f_{\mathrm{t}}([\mathrm{A\ video\ of\ Y_i}])\in \mathbb{R}^d$. Simultaneously, the video representation $v_i\in\mathbb{R}^d$ is obtained by averaging frame-level features, $r_{i, j}=f_{\mathrm{v}}(x_{i, j})\in\mathbb{R}^d$, as follows:

\begin{equation}
\label{eq;video_feature}
v_i = \frac{1}{T}\cdot \sum^T_{j=1}r_{i, j}\ .
\end{equation} 

\noindent The alignment between textual and video representations is quantified using cosine similarity:

\begin{equation}
\label{eq;cos_similarity}
\mathrm{cos}(v_i, t_i) = \frac{{v_i}^{\top} t_i}{\lVert v_i\rVert\cdot\lVert t_i\rVert}\ .
\end{equation}

\noindent Finally, the model is trained to maximize alignment between correct video-text pairs using a cross-entropy loss:

\begin{equation}
\label{eq;cross-entropy}
\mathcal{L}_{CE} = -\frac{1}{N}\sum_{i=1}^{N}\log{\frac{\exp(\mathrm{cos}(v_i, t_i)/\tau)}{\sum_{\mathcal{Y}}\exp(\mathrm{cos}(v_i, t_i)/\tau)}}\ .
\end{equation}

\begin{table*}[t] 
\caption{Details of evaluation datasets}
\begin{center}
\footnotesize
\begin{tabular}{C{2.8cm}C{0.9cm}C{2.8cm}C{1.5cm}C{0.7cm}C{1.2cm}C{0.7cm}}
\toprule
\multicolumn{1}{c}{{Dataset}} & {FPS} & {Source} & {Label Type} & {Class} & {Light} & {View}\\
\midrule
\multicolumn{1}{c}{UCF101 \cite{ucf101}} & 25-30 & Web-sourced & Verb+Noun & 101 & Well-lit & 3rd \\
\multicolumn{1}{c}{HMDB51 \cite{hmdb51}} & 30 & Web-sourced & Verb+Noun & 51 & Well-lit & 3rd \\
\multicolumn{1}{c}{SSv2 \cite{ssv2}} & 12 & Crowd-sourced & Verb & 174 & Well-lit & 1st \\
\multicolumn{1}{c}{ARID \cite{arid}} & 24-30 & Lab-collected & Verb & 11 & Dark & 3rd \\
\multicolumn{1}{c}{EK100$_{\mathrm{Verb}}$ \cite{epickitchens100}} & 50-60 & Participant-recorded & Verb & 97 & Well-lit & 1st \\
\bottomrule
\multicolumn{4}{l}{\footnotesize{$*$ \textit{FPS: Frames per second}}}
\end{tabular}
\label{tab:datasets}
\end{center}
\end{table*}

\subsection{Evaluation Datasets}
Table~\ref{tab:datasets} summarizes the five benchmark datasets used in our evaluation, selected to span a diverse range of visual variations. Our evaluation includes widely adopted datasets such as UCF101\footnote{https://www.crcv.ucf.edu/data/UCF101.php} \cite{ucf101}, HMDB51\footnote{https://serre.lab.brown.edu/hmdb51.html} \cite{hmdb51}, and SSv2\footnote{https://www.qualcomm.com/developer/software/something-something-v-2-dataset/downloads} \cite{ssv2}, as well as more challenging scenarios represented by ARID\footnote{https://xuyu0010.github.io/arid.html\#papers-and-download} \cite{arid} and EK100$_{\mathrm{Verb}}$\footnote{https://github.com/epic-kitchens/epic-kitchens-100-annotations/blob/master/README.md\#erratum} \cite{epickitchens100}. These benchmarks collectively allow for a comprehensive analysis of model robustness under various real-world conditions.
From the illumination perspective, ARID poses a significant challenge due to its low-light environments, which severely impair spatial perception. In terms of viewpoint, SSv2 and EK100$_{\mathrm{Verb}}$ introduce egocentric views, in contrast to the third-person viewpoints of other datasets. These egocentric datasets also tend to feature more complex motion dynamics, with SSv2 being notably motion-intensive. Lastly, EK100$_{\mathrm{Verb}}$ is semantically focused on kitchen-related activities, offering a narrower but more fine-grained action category distribution.
This diverse dataset composition ensures a well-rounded evaluation across visual challenges, facilitating a rigorous testbed for few-shot video action recognition.

\subsection{Quantification of Domain Shift}\label{section;diversity}
We quantify four key aspects of the datasets in feature space to investigate how domain shifts and dataset biases correlate with recognition performance. Inspired by CLIP scores \cite{clip-i}, we measure the distances between CLIP feature embeddings using cosine distance:

\begin{equation}
\label{eq;cos_distance}
\mathrm{d}(v_i, t_i)=1-\cos(v_i, t_i)\ .
\end{equation}

\noindent\textbf{Visual-Textual Misalignment.}
Due to the lack of detailed and informative textual descriptions, the visual context of videos is often not reflected in their corresponding labels. For example, the ARID dataset contains videos captured in low-light conditions, but the labels do not convey this information. This visual-textual gap can hinder the model’s ability to correctly align video content with the appropriate textual label. To assess this misalignment, we compute the discrepancy between the visual and textual representations for each video-text pair:

\begin{equation}
\label{eq;video-text misalignment}
\mathrm{CLIP}_{\mathrm{VTM}}=\frac{1}{N}\sum_{i=1}^N\mathrm{d}(v_i, t_i)\ .
\end{equation}

\noindent\textbf{Inter-Action Visual Distance.}
The distribution of visual features across different action categories can influence recognition difficulty. In particular, factors such as low lighting or egocentric viewpoints may cause visually distinct actions to appear similar in the feature space, making it harder for models to differentiate them. To capture this, we measure the pairwise distances between the centroids of visual features for each action class:

\begin{equation}
\label{eq;video_centroid}
\mu^c = \frac{1}{N^c}\sum_{i=1}^{N^c}v_i^c\ ,
\end{equation}

\begin{equation}
\label{eq;iavd}
\mathrm{CLIP}_{\mathrm{IAVD}}=\frac{2}{M(M-1)}\sum_{i<j}\mathrm{d}(\mu^i, \mu^j)\ .
\end{equation}

\noindent\textbf{Inter-Action Semantic Distance.}
Similar to visual distance, the semantic diversity of action labels can also impact recognition performance. On the textual side, we quantify this by measuring the pairwise distances between text embeddings of different action labels:

\begin{equation}
\label{eq;iatd}
\mathrm{CLIP}_{\mathrm{IASD}}=\frac{2}{M(M-1)}\sum_{i<j}\mathrm{d}(t^i, t^j)\ .
\end{equation}

\noindent\textbf{Motion Dynamics.}
Unlike static images, video data includes motion information along the temporal axis. To quantify the degree of motion within a video, we measure the average feature distance between adjacent frames:

\begin{equation}
\label{eq;motion_dynamics}
\mathrm{CLIP}_{\mathrm{MD}}=\frac{1}{N(T-1)}\sum_{i=1}^N\sum_{j=2}^T \mathrm{d}(r_{i, j}, r_{i, j-1})\ .
\end{equation}

\begin{figure*}[!t]
\centering
\includegraphics[width=\textwidth]{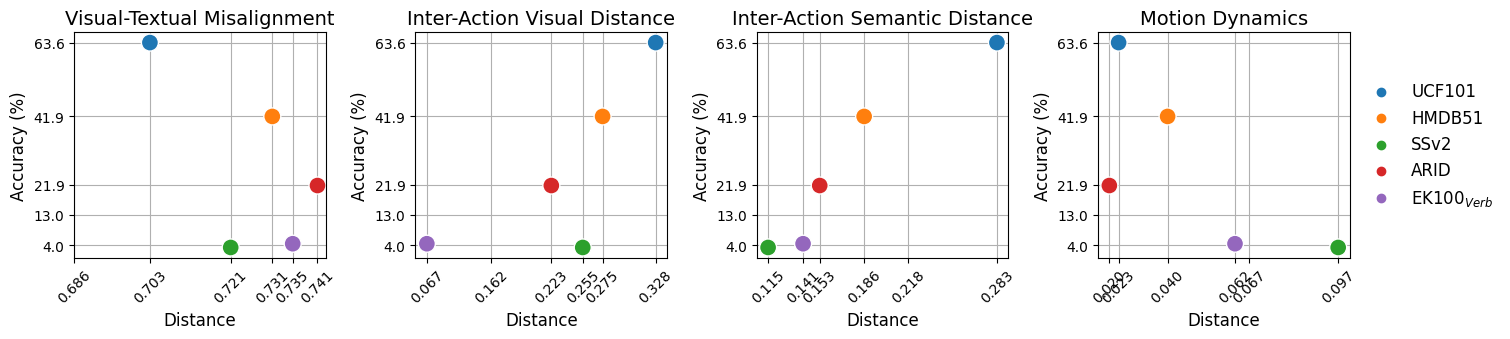}
\caption{Analysis of domain shift factors and their relationship with the zero-shot performance of Vanilla CLIP across video datasets. Each subfigure illustrates the correlation between CLIP’s top-1 accuracy and a specific domain property: (a) visual-textual misalignment, (b) inter-action visual distance, (c) inter-action semantic distance, and (d) motion dynamics.}
\label{fig;domain_shift}
\end{figure*}

\subsection{Analysis of Domain Shift}
We extracted embeddings using the CLIP with ViT-B/16 visual backbone from each evaluation dataset. For each video, eight frames were uniformly sampled from center-clipped video segments. Specifically, 32 frames were initially center-clipped for most datasets, while 16 frames were used for SSv2 and 64 frames for EK100$_\mathrm{Verb}$ to account for their respective frame rate differences.

Fig.~\ref{fig;domain_shift} illustrates how various domain shift factors correlate with action recognition performance. First, visually challenging datasets such as ARID and EK100$_\mathrm{Verb}$ exhibit large discrepancies between visual and textual embeddings, reflecting a visual shift from action semantics. This mismatch is largely attributed to their specific characteristics, including low illumination in ARID and egocentric viewpoints in EK100$_\mathrm{Verb}$, which contribute to significant performance degradation. Second, visual and semantic biases lead to tighter clustering of feature embeddings around certain regions in the feature space, which reduces inter-class separability and hinders accurate action discrimination. These biases are often more pronounced in datasets that are not web-sourced and instead originate from more constrained collection environments, as commonly seen in real-world target domains. Finally, large motion dynamics introduce additional temporal complexity, further complicating frame-wise representation learning and degrading recognition performance.

In this work, we primarily focus on addressing the challenges posed by visual shift and bias, which are pervasive in practical downstream scenarios where action categories and recording environments tend to be restricted.

\section{Methodology}

\subsection{EV-CLIP}
\begin{figure*}[t]
  \centering
   \includegraphics[width=\textwidth]{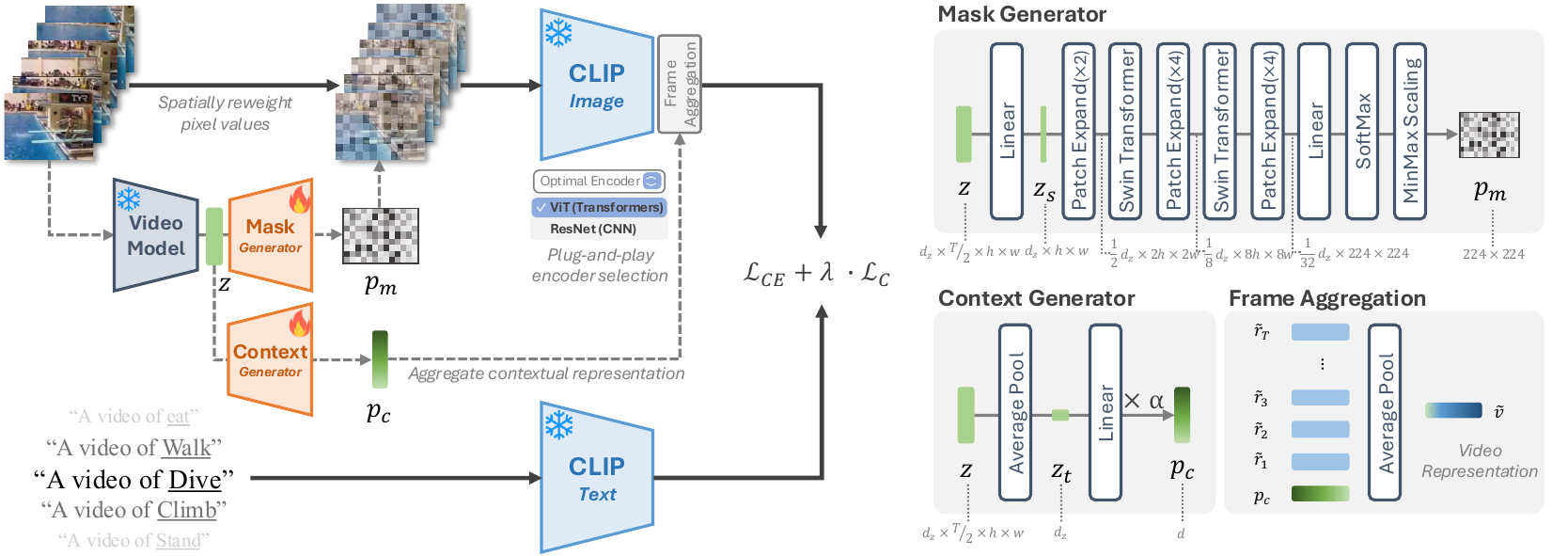}

   \caption{
   Overview of the EV-CLIP framework for video action recognition (left) and a detailed component breakdown (right). EV-CLIP enhances CLIP’s video understanding by introducing plug-and-play mask and context prompts with consistency loss while keeping CLIP frozen. The key components include: (a) Mask Generator, which produces mask prompts to reweight pixel values, highlighting action-relevant spatial features, and (b) Context Generator, which compresses video knowledge into context prompts, enabling temporal understanding. The video representation is obtained by merging frame features with the context prompt via parameter-free average pooling. Additionally, consistency loss reinforces a coherent action representation across frames, improving robustness to temporal variations.}
   \label{fig:overview}
\end{figure*}

EV-CLIP introduces mask and context prompts to align visual features with action semantics while keeping the image-pretrained CLIP model frozen, as depicted in Fig.~\ref{fig:overview}.  These prompts are generated by mask and context generators, which leverage latent features extracted from a pretrained video model (VM). Specifically, the latent feature $z\in \mathbb{R}^{d_z\times T⁄2\times h\times w}$ is obtained directly from a pretrained Omnivore model \cite{omni} without requiring additional training. This design allows EV-CLIP to enhance domain adaptation while maintaining trainable parameter efficiency. Further details on prompt generation mechanisms are provided in the following sections.

\noindent\textbf{Mask Prompt.}
The mask prompt reweights input frame pixels to highlight action regions, making them more visible and interpretable for CLIP. By leveraging video knowledge from the pretrained VM, the mask generator creates an input-customed mask that reinforces consistent action regions across frames, ensuring CLIP focuses on the target action while reducing background distractions.

To this end, the mask generator employs the decoder architecture of Swin-Unet \cite{swin-unet}. Initially, the latent feature $z$ is linearly projected into $z^s\in \mathbb{R}^{d_z\times h\times w}$, isolating spatial dimensions. The resolution is then upscaled using patch expanding layers and Swin transformer blocks. The patch expanding layer reshapes the feature with up-sampling resolutions using a linear layer. On the other hand, the Swin transformer blocks build hierarchical attention using a window-based multi-head self-attention (W-MSA) and a shifted W-MSA (SW-MSA). However, unlike the Swin-Unet, skip-connections are not feasible due to a dimensional mismatch with the VM. We therefore build a shallow architecture to prevent overfitting and enhance parameter efficiency. After the final patch expanding layer, a linear projection is applied to shape a single-channel mask. A softmax function is subsequently applied across pixels to highlight salient areas:

\begin{equation}
\label{eq6}
\tilde{z}^s_{i, j} = \frac{\exp({\hat{z}^s_{i, j}})}{\sum^H_{i=1}\sum^W_{j=1}\exp({\hat{z}^s_{i, j}})}\ ,
\end{equation} 
where $\hat{z}^s$ is the output of the final patch expanding layer. 
After softmax, the pixel weights can become excessively small due to the large number of pixels, which may inadvertently suppress important action-related cues. MinMax scaling rescales these outputs to $[0, 1]$, preventing the mask values from collapsing to near zero and preserving the visibility of critical cues:

\begin{equation}
\label{eq7}
p^m_{i, j} = \frac{\tilde{z}^s_{i, j}-\min(\tilde{z}^s)}{\max(\tilde{z}^s)-\min(\tilde{z}^s)}\ .
\end{equation} Finally, the mask prompt $p^m\in \mathbb{R}^{H\times W}$ is multiplied to all channels of every frame before the video frames are fed into CLIP, as described by:

\begin{equation}
\label{eq8}
\tilde{x}_{i, j} = {\{x_{i, j, k}\odot p^m\}}^C_{k=1}\ ,
\end{equation}

\begin{equation}
\label{eq9}
\tilde{r}_{i, j} = f_{\mathrm{v}}(\tilde{x}_{i, j})\ .
\end{equation}

\noindent\textbf{Context Prompt.}
The context prompt efficiently enhances video representation by providing contextual information to model temporal dependencies across frames. To enrich image-level frame features from CLIP, the context generator compresses the video knowledge into a prompt, providing a global temporal flow. Integrating this prompt with frame features enables CLIP to capture sequential variations, facilitating a more comprehensive video-level understanding.

The context generator begins by pooling the feature $z$ into $z^t\in\mathbb{R}^{d_z}$, retaining only the channel axis. Due to the dimensional discrepancy with frame features, $z_t$ is then linearly projected into $p^c\in\mathbb{R}^d$ to match the size suitable for integration, with multiplying a scale parameter $\alpha$. This scale parameter ensures that the prompt maintains its influence, even as the number of video frames increases. The final video representation is obtained by:

\begin{equation}
\label{eq10}
\tilde{v}_{i} = \frac{1}{T+1}( p^c+\sum^T_{j=1}\tilde{r}_{i, j})\ .
\end{equation}

\subsection{Consistency Loss}
CLIP processes each frame independently, making it challenging to maintain a consistent action representation across frames. This is particularly problematic when frame-to-frame variations introduce unnecessary diversity, potentially causing CLIP to misinterpret the action. To mitigate this, we introduce consistency loss to reinforce coherent representations across frames, ensuring that even visually varied frames contribute to a consistent action representation. By increasing the similarity across frame features, this loss reduces excessive variations that could distract the model from the main action, allowing CLIP to focus on action-relevant features rather than frame-specific differences.

For this, inter-frame similarity is scaled into $[0,1]$ and penalized in log-scale, where $J$ represents a matrix of ones:

\begin{equation}
\label{eq11}
s=\frac{1}{2}(\cos(\tilde{r}_i,\tilde{r}_i) + J_{T\times T})\ ,
\end{equation}

\begin{equation}
\label{eq12}
\mathcal{L}_{C}=-\frac{2}{T(T-1)}\sum_{i<j}\log(s_{i, j})\ .
\end{equation}

\noindent This loss is integrated into the cross-entropy loss using a hyper-parameter $\lambda$, optimized for each target dataset:

\begin{equation}
\label{eq13}
\mathcal{L}=\mathcal{L}_{CE}+\lambda\cdot\mathcal{L}_{C}\ .
\end{equation}

\section{Experiments}
\subsection{Experimental Setup}
We use ViT-B/16 as the visual encoder in CLIP, paired with Omnivore-small as the video model, without additional post-pretraining. Video clips are sampled with 8 frames, using randomly selected starting frames for training and center clipping for testing. Frames are resized to $256 \times 340$ and cropped to $224 \times 224$, using random cropping for training and central cropping for testing. 
The $\lambda$ values in the consistency loss were selected from the set $\{0, 0.01, 0.1, 1.0, 10.0\}$ and optimized individually for each dataset. As a result, 1.0 was chosen for EGTEA, 0.1 for ARID, HMDB51, and EK100$_\mathrm{Verb}$, 0.01 for SSv2, and 0 for UCF101. All experiments were conducted on a single NVIDIA A100 80G GPU.

\subsection{Comparative Methods.}
Before presenting the experimental results, we first introduce efficient CLIP-based video learners that are used for comparison in this paper. For the results that were not reported in their original papers, we re-implemented and evaluated them under our experimental setup. Notably, ViLT-CLIP, one of the recent approaches, is compared only on UCF101, HMDB51, and SSv2, as its code was not publicly released.

\noindent\textbf{A5.} Ju et al. \cite{a5} is one of the early studies that explored efficient tuning of CLIP for video action recognition. Their approach aggregates frame-level features extracted from CLIP’s image encoder using a 2-layer transformer, while adopting trainable text prompts for label inputs, following CoOp \cite{coop}. 

\noindent\textbf{ST-Adapter.} Pan et al. \cite{st-adapter} proposed a spatio-temporal adapter (ST-Adapter) for transferring image-pretrained knowledge to video domains efficiently. ST-Adapter introduces lightweight depthwise 3D convolutional layers \cite{x3d} between transformer blocks, enabling the model to learn spatio-temporal representations. In this study, we adopt ST-Adapter within the CLIP visual encoder, and perform action recognition by probing the video features against text embeddings using cosine similarity.

\noindent\textbf{AIM.} Yang et al. \cite{aim} proposed adapting image models for efficient video action recognition (AIM), which adapts pretrained image transformers to video domains. AIM introduces joint adaptation by inserting three types of adapters between Transformer blocks: spatial adapters for spatial feature refinement, temporal adapters for capturing temporal dependencies, and parallel adapters for additional feature transformation. In this paper, we adopt AIM within the CLIP visual encoder and perform action recognition by probing video features against text embeddings using cosine similarity.

\noindent\textbf{ViLT-CLIP.} Wang et al. \cite{vilt-clip} proposed ViLT-CLIP, a video and language tuning CLIP for video understanding tasks. ViLT-CLIP introduces trainable prompt tokens into every transformer block of both the visual and textual encoders. These prompt tokens are shared across modalities by switching them between the visual and textual branches through linear projections, following a design similar to MaPLe \cite{maple} and Ta-Adapter \cite{taadapter}. Additionally, it incorporates hand-crafted prompts on the textual side to mitigate the forgetting of scenario-specific knowledge. Due to the lack of publicly available code and reproducibility challenges, we report their performance only on UCF101, HMDB51, and SSv2, based on the results provided in their original paper.

\noindent\textbf{EZ-CLIP.} Ahmad et al. \cite{ez-clip} proposed efficient zero-shot CLIP (EZ-CLIP) for video action recognition. EZ-CLIP introduces trainable prompt tokens into the visual branch, which are aggregated using a multi-head attention mechanism to capture temporal dependencies. Furthermore, EZ-CLIP incorporates lightweight adapter layers within the transformer blocks of the visual and textual encoders, facilitating better adaptation. To enrich and semantic representation of action categories, EZ-CLIP additionally employs GPT-3.5 to automatically generate informative action descriptions. Motion loss further brings dynamics between frame features. Following their approach, we similarly generate action descriptions using either GPT-3.5 or GPT-4.0 for ARID, EGTEA, and EK100$_{\mathrm{Verb}}$ datasets in the re-implementation of EZ-CLIP.

\begin{figure*}[t]
  \centering
   \includegraphics[width=0.9\textwidth]{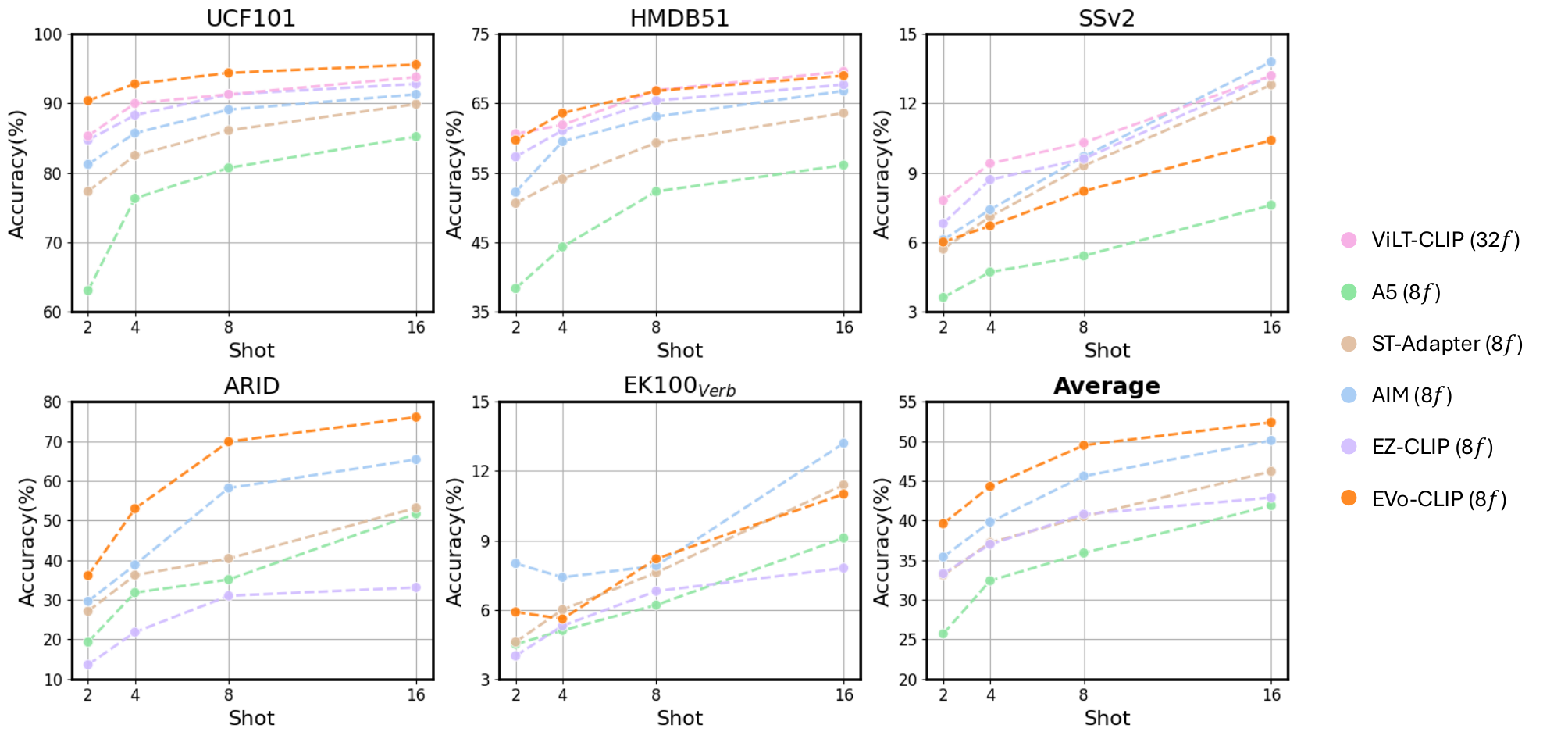}

   \caption{
   Performance comparison across five benchmark datasets from two-shot to 16-shot settings using the ViT-B/16 backbone, with the average plot summarizing overall adaptability.
   }
   \label{fig:petm}
\end{figure*}

\begin{figure*}[t]
  \centering
   \includegraphics[width=\textwidth]{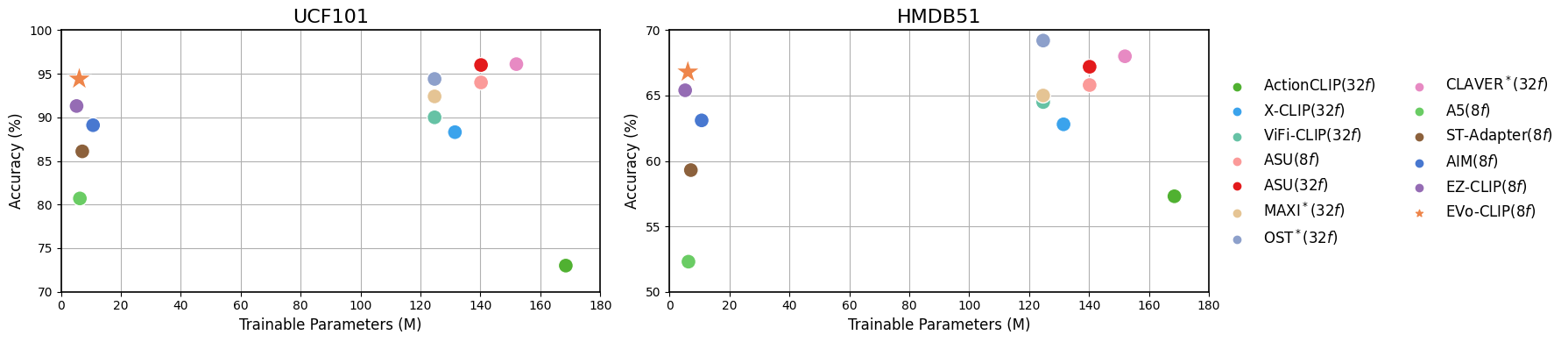}

   \caption{
   Training efficiency comparison in terms of trainable parameters versus performance on UCF101 and HMDB51 under eight-shot settings. $^*$ indicates methods with additional post-pretraining on the Kinetics-400 dataset \cite{kinetics}.
   }
   \label{fig:efficiency}
\end{figure*}

\subsection{Comparison with PETMs.}
Fig.~\ref{fig:petm} illustrates the few-shot performance of efficient CLIP-based learners across five benchmark datasets, ranging from two-shot to 16-shot settings. 

EV-CLIP demonstrates the strongest overall adaptability, achieving the highest average performance among all PETMs. In particular, it significantly outperforms all baselines in low-light conditions and shows robustness for egocentric videos. While it still faces challenges under complex motion dynamics, it demonstrates robust spatial perception, which is essential for effective temporal understanding.

Specifically, on web-sourced datasets, EV-CLIP achieves state-of-the-art performance on UCF101 and remains competitive on HMDB51, despite using only eight frames compared to ViLT-CLIP, which processes 32-frame sequences. A particularly notable advantage is observed on ARID, where EV-CLIP significantly outperforms all baselines. This dataset, characterized by low-light conditions, poses a serious challenge for conventional models. We attribute EV-CLIP’s success to its mask prompt, which reweights pixel intensities to amplify action-relevant regions and improve visual saliency under dark conditions. In contrast, other methods directly process raw pixel values, which remain largely indistinct in poor lighting.

However, EV-CLIP still faces challenges on SSv2 and EK100${\mathrm{Verb}}$ due to their complex motion dynamics. While the lightweight high-level temporal aggregation employed by the context prompt proves effective in many scenarios, our analysis reveals that capturing intricate motion patterns often requires deeper, layer-wise temporal modeling from low to high-level features. This is exemplified by methods such as AIM and ST-Adapter, whose more elaborate temporal modeling strategies demonstrate stronger performance on these datasets. Nevertheless, such approaches often rely on tightly coupled architectural modifications, which compromise modularity and scalability, making them less ideal for general-purpose adaptation. Another noteworthy observation is the performance of EZ-CLIP. Although it performs competitively on SSv2, it struggles on EK100${\mathrm{Verb}}$, suggesting that visual and semantic biases undermine its adaptability. In contrast, EV-CLIP exhibits robustness across these challenging domains and achieves the best performance at the 8-shot setting on EK100$_{\mathrm{Verb}}$, demonstrating that even lightweight temporal modeling can yield strong results when paired with effective spatial perception mechanisms.

Despite these challenges, the averaged performance across all datasets confirms that EV-CLIP provides the best trade-off between efficiency and adaptability. It not only performs robustly on web-sourced benchmarks but also delivers significant gains in visually degraded environments, making it a strong candidate for real-world deployment under domain shift and data scarcity.

\begin{table*}[t!]
\caption{
Computational costs under UCF101 settings
}
\centering
\begin{tabular}{C{2.0cm}C{2.0cm}C{2.0cm}C{2.0cm}C{3.0cm}}
\toprule
\multicolumn{1}{c}{Backbone} & \multicolumn{1}{c}{Method} & FLOPs(G) & Throughput & Trainable Params (M) \\
\midrule
\multirow{6}{*}{ViT-B/16} & Vanilla CLIP & 429.0 & 223.5/sec & - \\
 & A5 & 429.1 & 70.9/sec & 6.3 \\
 & ST-Adapter & 440.2 & 62.6/sec & 7.1 \\
 & AIM & 490.5 & 45.2/sec & 10.7 \\
 & EZ-CLIP & 447.3 & 61.6/sec & 5.2 \\
 & \cellcolor{Gray}\textbf{EV-CLIP} &\cellcolor{Gray}468.7 & \cellcolor{Gray}85.2/sec & \cellcolor{Gray}6.1 \\
\midrule
\multirow{6}{*}{ViT-L/14} & Vanilla CLIP & 1,284.0 & 56.6/sec & - \\
 & A5 & 1,284.1 & 17.0/sec & 14.2 \\
 & ST-Adapter & 1,309.8 & 14.8/sec & 12.6 \\
 & AIM & 1,568.8 & 10.3/sec & 37.8 \\
 & EZ-CLIP & 1,339.0 & 13.1/sec & 10.1 \\
 & \cellcolor{Gray}\textbf{EV-CLIP} & \cellcolor{Gray}1,323.6 & \cellcolor{Gray}18.1/sec & \cellcolor{Gray}6.3 \\
\midrule
\multirow{3}{*}{ResNet50} & Vanilla CLIP & 343.1 & 415.3/sec & - \\
& A5 & 343.3 & 75.4/sec & 25.2 \\
 & \cellcolor{Gray}\textbf{EV-CLIP} & \cellcolor{Gray}382.8 & \cellcolor{Gray}102.6/sec & \cellcolor{Gray}6.4 \\

\bottomrule
\end{tabular}
\label{tab:computational_costs}
\end{table*}

\subsection{Computational Efficieny.}
Fig.~\ref{fig:efficiency} compares the trainable parameter counts and the corresponding performance of CLIP-based action learners under eight-shot settings on the widely adopted UCF101 and HMDB51 benchmarks. As expected, full fine-tuning approaches exhibit significantly larger parameter counts due to updates across all layers of the CLIP encoder. In contrast, PETMs retain the pretrained weights, yielding a drastically reduced number of trainable parameters. EV-CLIP achieves superior performance among PETMs, even being competitive with finetuning methods. This highlights the effectiveness of EV-CLIP’s modular visual prompting, which enhances adaptability to new domains while maintaining training efficiency.

Table~\ref{tab:computational_costs} further breaks down the computational cost of each PETM by examining FLOPs, throughput, and trainable parameter scalability across different backbone choices. While EV-CLIP introduces a modest increase in FLOPs due to the video-based prompt generators, it achieves notably higher throughput compared to other PETMs. This efficiency is crucial for real-time or large-scale deployment.

We attribute the strong throughput performance to several design factors. First, unlike methods that insert adapters or prompts throughout each transformer layer, thereby incurring latency proportional to model depth, EV-CLIP employs a modular and non-intrusive design. Its visual prompts are computed externally and injected only once, avoiding expensive layer-wise processing. Additionally, our Swin transformer–based prompt generation efficiently captures spatiotemporal patterns, offering a favorable trade-off between representation capacity and computational load.

Finally, EV-CLIP’s backbone-agnostic nature allows it to flexibly adapt to lightweight or heavy encoders depending on deployment needs. This flexibility enables users to scale computational cost independently from adaptation capacity, a property not afforded by many prior PETMs that rely on deep integration with specific transformer architectures or high dependency on backbone scales.

\begin{table*}[t!]
\caption{
Comparison with temporal modules under eight-shot settings
}
\centering
\begin{tabular}{ccC{1.5cm}C{1.5cm}C{1.5cm}C{3.0cm}}
\toprule
Aggregation Method & \multicolumn{1}{c}{Layers} & UCF101 & HMDB51 & ARID & Trainable Params (M)\\
\midrule
LSTM & 1 & 82.4 / \underline{96.6} & \underline{53.5} / \underline{82.7} & 51.7 / 85.4 & 2.1 \\
LSTM & 2 & 77.9 / 94.9 & 46.7 / 75.8 & 48.8 / 85.2 & 4.2 \\
Transformer & 1 & 81.8 / 96.0 & 52.7 / 81.2 & 50.3 / 85.5 & 3.2 \\
Transformer & 2 & \underline{82.8} / 96.0 & 52.5 / 82.2 & \underline{53.3} / \underline{86.5} & 6.3 \\
\rowcolor{Gray}
Context Prompt & - & \textbf{94.4} / \textbf{99.2} & \textbf{66.8} / \textbf{87.9} & \textbf{69.9} / \textbf{95.7} & 0.4 \\
\bottomrule
\end{tabular}
\label{tab:temporal_aggregation}
\end{table*}

\subsection{Comparison with Temporal Modules}
EV-CLIP utilizes a context prompt for efficient temporal aggregation, offering a parameter-efficient alternative to sequential layers like LSTM and Transformer used in previous studies \cite{a5, actionclip}. As shown in Table~\ref{tab:temporal_aggregation}, LSTM and Transformer layers require at least five times more parameters than the context prompt but still underperform. Furthermore, increasing layers in LSTM or Transformer models yields minimal gains and can even reduce performance in LSTM. These findings indicate that the context prompt is both parameter-efficient and effective under limited supervision, outperforming complicated sequential models.

\subsection{Ablation Study}
To further understand the effectiveness of our design choices, we conduct ablation studies on HMDB51, SSv2, and ARID. We report both top-1 and top-5 accuracies to offer a comprehensive analysis of model performance under each variant.

\begin{table*}[t!]
\caption{
Ablation study on CLIP visual backbones
}
\setlength{\tabcolsep}{2pt}
\centering
\begin{tabular}{C{3.0cm}C{1.0cm}C{1.0cm}C{1.0cm}C{1.0cm}C{1.0cm}C{1.0cm}C{1.0cm}C{1.0cm}C{1.0cm}C{1.0cm}C{1.0cm}C{1.0cm}C{2.0cm}}
\toprule
\multicolumn{1}{c}{\multirow{2}{*}{\begin{tabular}{cc}
     Visual  \\
     Backbone 
\end{tabular} }} & \multicolumn{4}{c}{HMDB51} & \multicolumn{4}{c}{SSv2} & \multicolumn{4}{c}{ARID} & \multicolumn{1}{c}{\multirow{2}{*}{\begin{tabular}{cc}
     Total  \\
     Params (M) 
\end{tabular} }} \\
\cmidrule(lr{0.5em}){2-5}\cmidrule(lr{0.5em}){6-9}\cmidrule(lr{0.5em}){10-13}
\multicolumn{1}{c}{} & \textit{K=2} & \textit{K=4} & \textit{K=8} & \multicolumn{1}{c}{\textit{K=16}} & \textit{K=2} & \textit{K=4} & \textit{K=8} & \multicolumn{1}{c}{\textit{K=16}} & \textit{K=2} & \textit{K=4} & \textit{K=8} & \multicolumn{1}{c}{\textit{K=16}} & \\
\midrule
\multicolumn{1}{c}{ResNet50} & 51.4 & 57.5 & 63.5 & \multicolumn{1}{c}{66.1} & 5.1 & \underline{6.5} & 7.8 & \multicolumn{1}{c}{9.6} & \underline{40.1} & \underline{59.7} & 69.6 & \multicolumn{1}{c}{\textbf{77.4}} & 159.1 \\
\multicolumn{1}{c}{ResNet101} & 50.9 & 58.1 & 64.6 & \multicolumn{1}{c}{64.8} & 4.9 & 6.2 & 7.2 & \multicolumn{1}{c}{8.8} & \textbf{44.3} & \textbf{60.8} & \underline{70.3} & \multicolumn{1}{c}{76.8} & 176.3 \\
\multicolumn{1}{c}{ViT-B/32} & 57.5 & 62.0 & 65.3 & \multicolumn{1}{c}{67.4} & \underline{5.6} & 5.9 & 7.7 & \multicolumn{1}{c}{9.6} & 39.8 & 55.1 & 69.1 & \multicolumn{1}{c}{\underline{77.2}} & 207.9 \\
\multicolumn{1}{c}{ViT-B/16}  & \underline{59.7} & \underline{63.6} & \underline{66.8} & \multicolumn{1}{c}{\underline{69.0}} & \textbf{6.0} & \textbf{6.7} & \textbf{8.2} & \multicolumn{1}{c}{\textbf{10.4}} & 36.1 & 53.0 & 69.9 & \multicolumn{1}{c}{76.1}  & 206.3 \\
\multicolumn{1}{c}{ViT-L/14} & \textbf{62.4} & \textbf{64.4} & \textbf{69.5} & \multicolumn{1}{c}{\textbf{70.6}} & 4.9 & 6.4 & \underline{7.9} & \multicolumn{1}{c}{\underline{9.9}} & 36.5 & 52.8 & \textbf{71.0} & \multicolumn{1}{c}{74.9}  & 484.5 \\
\bottomrule

\multicolumn{14}{l}{\footnotesize{$*$ \textit{Bold indicates the highest accuracy, while underlined denotes the second highest.}}}
\end{tabular}%
\label{tab:table_clip_backbone}
\end{table*}

\noindent\textbf{Effect of CLIP Backbone Choice.}
Our modular prompting strategy enables backbone-agnostic adaptation, offering practical flexibility in balancing performance and computational efficiency. Table~\ref{tab:table_clip_backbone} reports the few-shot performance of EV-CLIP across various CLIP visual backbones, alongside their total parameter counts.

Notably, EV-CLIP delivers strong performance even when equipped with the lightweight ResNet50, significantly reducing computational costs without compromising accuracy. Interestingly, larger backbones such as ViT-L/14 do not consistently yield the best performance, particularly on visually challenging datasets like SSv2 and ARID, where lighter backbones sometimes outperform them. This trend suggests that increased model size does not always correlate with improved domain adaptability, especially under severe visual shift or illumination degradation.

These findings underscore the practical value of EV-CLIP’s modularity. Its ability to maintain high adaptability with lighter backbones allows users to select optimal performance-efficiency trade-offs based on deployment constraints, making it highly suitable for real-world applications.

\begin{table}[t!]
\caption{
Ablation study on video model scale}
\centering
\begin{tabular}{ccccc}
\toprule
\multicolumn{1}{c}{
\makecell{VM\\Scale}} & HMDB51 & SSv2 & \multicolumn{1}{c}{ARID} & \makecell{Trainable\\Params(M)} \\ 
\midrule
Tiny & 63.8 / 86.4 & 7.7 / 21.8 & 60.9 / 92.9 & 6.1 \\
\rowcolor{Gray}
Small & \textbf{66.8} / \textbf{87.9} & \textbf{8.2} / \textbf{22.8} & \textbf{69.9} / \textbf{95.7} & 6.1 \\
Base & \underline{66.3} / \underline{87.8} & \underline{7.8} / \underline{21.9} & \textbf{69.9} / \underline{95.5} & 10.6 \\
\bottomrule
\end{tabular}
\label{tab:vm_choice}
\end{table}

\noindent\textbf{Effect of Video Model Scale.}
Omnivore-small model is employed in EV-CLIP to generate mask and context prompts that bridge modality and domain gaps. Table~\ref{tab:vm_choice} reveals that a small-scale video model is sufficient to produce effective prompts for adaptation.

While a base-sized video model introduces additional trainable parameters and computational overhead, mainly due to its higher-dimensional latent representations, it does not consistently improve performance. In HMDB51 and SSv2, the larger model even underperforms, likely due to overfitting or overrepresentation in the prompt space. Conversely, a tiny model maintains competitive accuracy, reducing computational cost.

These results highlight that the video model in EV-CLIP functions primarily as a supportive module for prompt generation, rather than as the main driver of action recognition. Therefore, we adopt the Omnivore-small model as our default video model, striking a favorable balance between efficiency and adaptation performance.

\begin{table}[t!]
\caption{
Ablation study on the impact of mask and context prompts
}
\centering
\begin{tabular}{ccccc}
\toprule
Mask & \multicolumn{1}{c}{Context} & HMDB51 & SSv2 & ARID \\
\midrule
- & - & 41.8 / 72.2 & 3.4 / 11.2 & 21.6 / 61.8\\
\checkmark & - & 59.6 / 81.0 & 6.0 / 16.7 & \underline{65.4} / 91.4 \\
- & \checkmark & \underline{66.1} / \underline{87.4} & \textbf{8.5} / \textbf{23.0} & 52.1 / \underline{92.0} \\
\rowcolor{Gray}
\checkmark & \checkmark & \textbf{66.8} / \textbf{87.9} & \underline{8.2} / \underline{22.8} & \textbf{69.9} / \textbf{95.7} \\ 
\bottomrule
\end{tabular}
\label{tab:ablation_mask_vs_context}
\end{table}

\noindent\textbf{Mask vs. Context Prompts.}
Table~\ref{tab:ablation_mask_vs_context} compares individual and combined contributions of the mask and context prompts. When evaluated separately, the context prompt proves more effective on well-lit datasets such as HMDB51 and SSv2, underscoring its effectiveness in temporal modeling across frames. Meanwhile, the mask prompt demonstrates pronounced benefits on the visually challenging ARID dataset, where it mitigates the adverse effects of low-light conditions by enhancing visibility.

Importantly, the combination of both prompts consistently yields promising performance across datasets. This synergy suggests that spatial enhancement and temporal context modeling complement each other, making the joint design crucial for robust action recognition under diverse visual conditions.

\begin{figure}[t]
  \centering
   \includegraphics[width=\columnwidth]{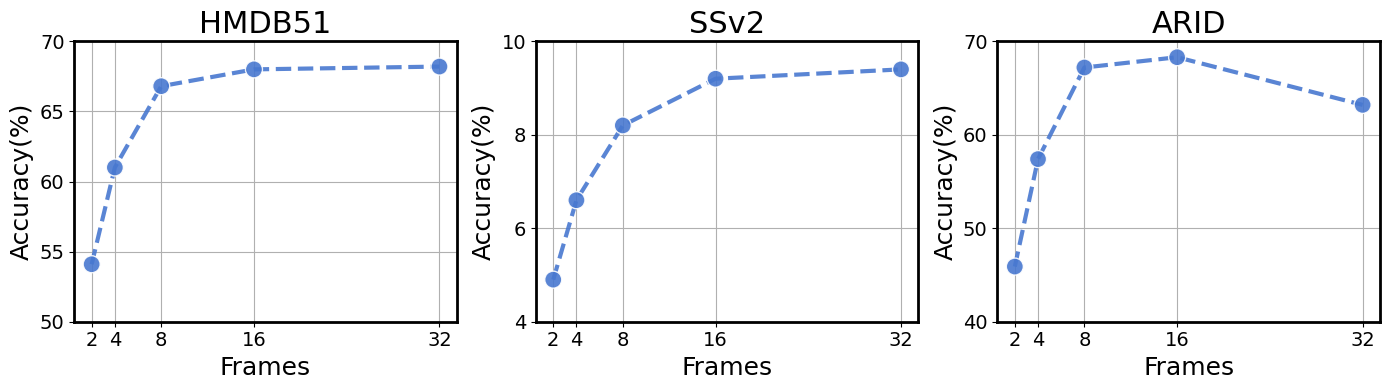}

   \caption{
   Impact of frame length on performance across all evaluation datasets.
   }
   \label{fig:frames}
\end{figure}

\noindent\textbf{Effect of Frame Numbers.}
Fig.~\ref{fig:frames} presents the impact of varying the number of input frames on recognition performance. As expected, increasing the number of frames generally improves performance by offering richer temporal context. However, the performance gain plateaus beyond 16 frames, and in the case of ARID, using 32 frames even leads to a slight performance drop. This indicates that excessive frames can introduce visual redundancy or irrelevant motion, which may obscure salient cues and degrade model focus. 

Moreover, increasing the number of input frames leads to higher computational costs in both training and inference. To strike a balance between performance and efficiency, we identify the 8-frame setting as the optimal trade-off. It consistently provides strong performance while avoiding diminishing returns and additional resource demands associated with longer sequences.

\begin{figure*}[t]
  \centering
   \includegraphics[width=\textwidth]{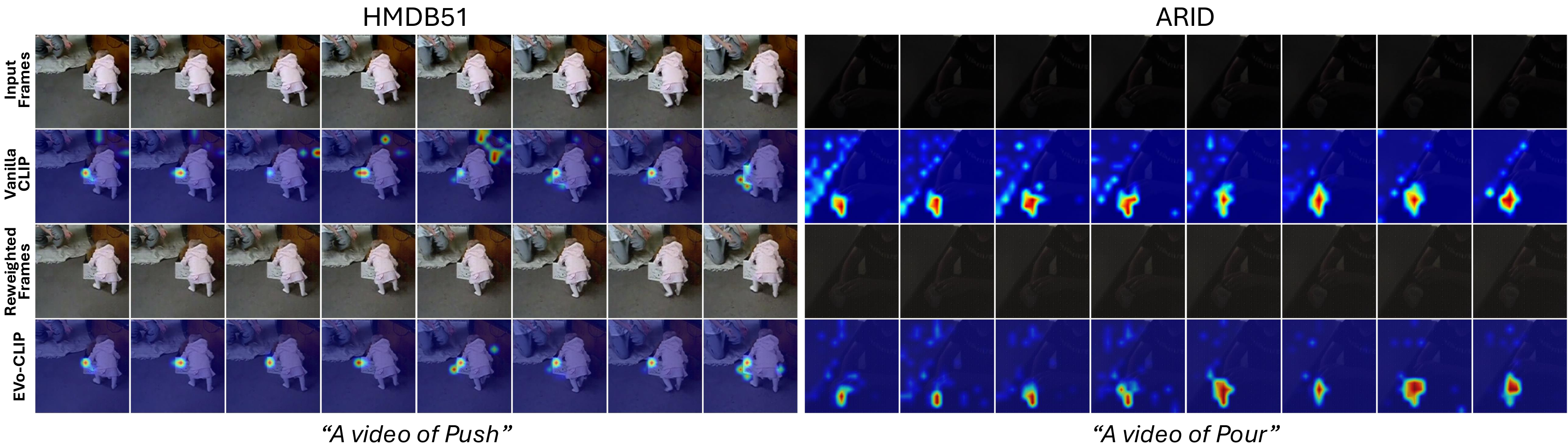}
   \caption{
   Visualization of frames and attention heatmaps comparing Vanilla CLIP and EVo-CLIP on two examples from HMDB51 and ARID evaluation sets.
   }
   \label{fig:gradcam}
\end{figure*}

\subsection{Visualization}

\noindent\textbf{Attention Heatmap.}
To provide interpretable insights for performance enhancement, Grad-CAM \cite{grad-cam} visualizations are presented in Fig.~\ref{fig:gradcam}. Vanilla CLIP, which relies on image-level knowledge, exhibits inconsistent attention across frames, often focusing on irrelevant areas. This issue is particularly evident in ARID, where the model further attends to the wall in the background, which has no interaction with the pouring hand. In contrast, EV-CLIP yields consistent attention only to the action-relevant regions, effectively filtering out irrelevant distractions. Moreover, in low-light conditions,  the mask prompt’s pixel reweighting mechanism enhances visibility by adjusting pixel values toward a brighter condition. These qualitative results highlight EV-CLIP’s effectiveness as an efficient yet powerful solution for domain adaptation in video action recognition, ensuring improved interpretability and robustness across diverse environments.

\begin{figure*}[t]
  \centering
   \includegraphics[width=0.8\textwidth]{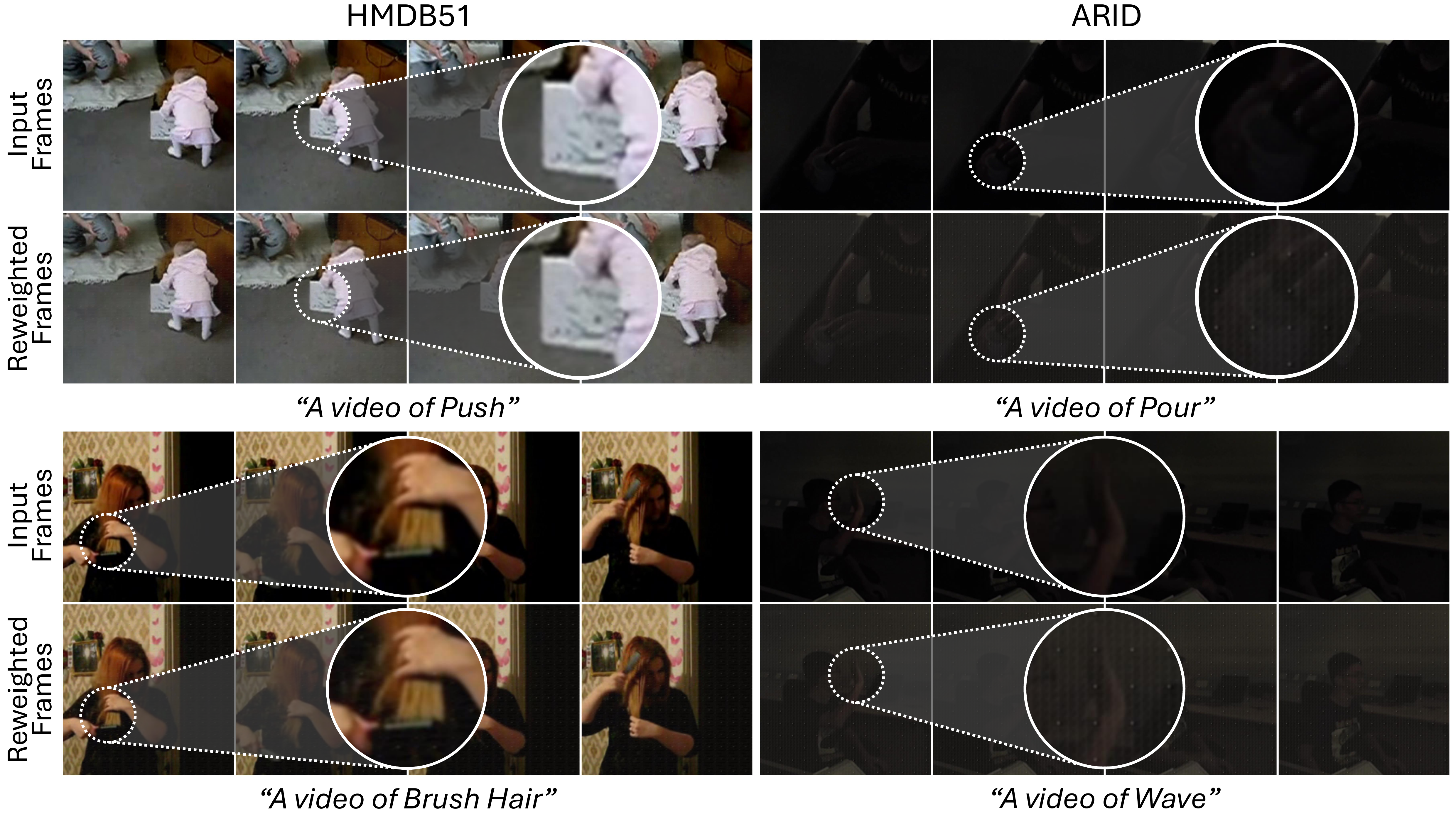}

   \caption{
   Visualization of original and reweighted frames on HMDB51 and ARID validation sets.
   }
   \label{fig:mask}
\end{figure*}

\noindent\textbf{Reweighted Input Frames.}
To further analyze the effects of the mask prompt, we examine the visual modifications in reweighted input frames. Fig.~\ref{fig:mask} presents both the original input frames and their reweighted versions from HMDB51 and ARID.

For HMDB51, where visibility conditions are clear, the mask prompt introduces minimal modifications, as strong reweighting is unnecessary. In contrast, ARID samples become noticeably brighter, effectively mitigating the low-light conditions and improving overall visibility.

Additionally, we observe a dotted pattern appearing in both datasets after reweighting. This phenomenon likely results from the MinMax scaling operation, which prevents the mask from darkening the entire frame. However, this process assigns near-zero weights to certain pixels, and when these weights are applied to normalized pixel values, they produce zero-valued pixels. Upon denormalization, these pixels appear as gray spots, forming the observed pattern. The model may intentionally distribute these pixels to minimize their impact on visual recognizability.

\section{Discussion}
In this section, we discuss the current limitations and design choices of EV-CLIP, along with its practical implications for real-world deployment.

EV-CLIP enhances spatial perception by emphasizing salient regions through a mask generated by a compact Swin-Unet inspired module, designed specifically for robust domain adaptation. Although our previous CNN-based mask generator \cite{watermark} required fewer trainable parameters, EV-CLIP provides richer contextual awareness and consistently yields superior performance under visually degraded conditions, demonstrating a favorable balance between modeling capacity and efficiency. On the other hand, EV-CLIP models temporal dynamics through a context prompt, which is aggregated with frame features at high-level to bridge the modality gap between static image inputs and dynamic video content. While this design introduces a lightweight temporal modeling without modifying internal CLIP layers, we observe limitations when handling videos with high motion dynamics. Notably, A5, which also performs temporal aggregation at high levels, demonstrates similar challenges. 
In contrast, prior works that explicitly model temporal alignment across frames, such as approaches that introduce temporal modules (\textit{e.g.}, adapters \cite{st-adapter, aim} or trainable global tokens \cite{vita-clip}) throughout multiple layers of the encoder, often demonstrate stronger performance on motion-intensive datasets. By enabling temporal information exchange from early to late stages of feature extraction, these methods facilitate more fine-grained modeling of frame-wise variations and motion dependencies. 
These findings suggest that layer-wise temporal modeling, spanning low to high-level representations, is essential for capturing fine-grained temporal variability. However, such designs often require intrusive architectural modifications and are tightly coupled to transformer-based backbones, resulting in increased parameter counts and reduced architectural flexibility. By contrast, EV-CLIP offers a backbone-agnostic solution. Its plug-and-play visual prompts enable effective temporal adaptation without introducing structural dependencies, thereby maintaining both efficiency and generalizability across different encoders, including lightweight CNNs. This design makes EV-CLIP more practical for real-world deployments where computational resources are limited.

While EV-CLIP focuses on visual-side adaptation, textual semantics also play a critical role in action recognition, particularly under domain shifts. Previous works such as EZ-CLIP, ViLT-CLIP, and A5 have attempted joint tuning of both visual and textual encoders to bridge modality gaps. We also explored similar strategies, adapting CoOp-style prompt tuning \cite{coop} for CLIP’s text encoder. However, our experiments revealed that simultaneous tuning of both modalities often leads to performance degradation. We hypothesize this is due to semantic instability in textual label embeddings, which disrupts alignment and weakens optimization stability. Supporting this, AA-CLIP \cite{aa-clip}, designed for anomaly detection, proposes a two-stage textual adaptation: first amplifying inter-class distances in the text space, then aligning visual features accordingly. These findings suggest that decoupling the adaptation of visual and textual modalities can lead to more stable and effective transfer. EV-CLIP adheres to this principle, adapting only the visual side. Despite this constraint, it demonstrates strong generalization across visually diverse domains, without incurring the computational and optimization costs associated with textual tuning.

\section{Conclusion}
This paper introduces EV-CLIP, a modular and efficient framework for adapting CLIP to video action recognition under visual challenges. In contrast to prior methods that depend on extensive fine-tuning or internal architectural modifications, EV-CLIP employs two lightweight visual prompts: mask prompts, which enhance spatial focus by reweighting frames at the pixel level, and context prompts, which efficiently model global temporal dynamics by compressing frame-wise features. This design ensures parameter-efficient adaptation without altering the CLIP backbone, maintaining compatibility across both CNN and transformer-based encoders. To enable a comprehensive evaluation, we curate and analyze five benchmark datasets encompassing various visual challenges such as egocentric views and low-light conditions, demonstrating that multiple visual and semantic factors influence CLIP’s recognizability. Experimental results show that EV-CLIP achieves state-of-the-art performance among efficient adaptation methods. Notably, it retains strong accuracy even when paired with lightweight backbones, significantly reducing computational overhead and inference latency. These findings position EV-CLIP as a practical and scalable solution for real-world video understanding tasks in data- and resource-constrained settings.

\bibliographystyle{IEEEtran}
\bibliography{main}

\vfill

\end{document}